\documentclass[sigconf]{acmart}

\AtBeginDocument{%
  }

 \renewcommand\footnotetextcopyrightpermission[1]{}

\usepackage{pifont}
\usepackage[table]{xcolor}
\usepackage{booktabs}
\usepackage{multirow}
\usepackage{array}
\usepackage{graphicx}
\usepackage{makecell}
\usepackage{tcolorbox}
\usepackage{caption}
\usepackage{enumitem}

\usepackage{titlesec}

\titlespacing{\section}{0pt}{*0.5}{*0.5}     % {left}{before-sep}{after-sep}
\titlespacing{\subsection}{0pt}{*0.5}{*0.5}

\begin{document}

\title{Open3D-VQA: A Benchmark for Embodied Spatial Reasoning with Multimodal Large Language Model in Open Space
}

\author{
 \textbf{Weichen Zhang\textsuperscript{1,2}*},
 \textbf{Zile Zhou\textsuperscript{1}*},
 \textbf{Xin Zeng\textsuperscript{3}*},
 \textbf{Xuchen Liu\textsuperscript{2}},
 \textbf{Jianjie Fang\textsuperscript{1}},
 \textbf{Chen Gao\textsuperscript{1$\ddagger$}},
 \\
 \textbf{Yong Li\textsuperscript{1}},
 \textbf{Jinqiang Cui\textsuperscript{2}},
 \textbf{Xinlei Chen\textsuperscript{1$\ddagger$}},
 \textbf{Xiao-Ping Zhang\textsuperscript{1}}
\\
 \textsuperscript{1}Tsinghua University, 
 \textsuperscript{2}Pengcheng Laboratory,
 \textsuperscript{3}Sun Yat-sen University
 \\
 \textsuperscript{}*Equal Contribution,
 \textsuperscript{$\ddagger$}Corresponding Author
}

% \author{
% Weichen Zhang$^{1,2*}$, 
% Zile Zhou$^{1*}$, 
% Xin Zeng$^{3*}$, 
% Xuchen Liu$^2$, 
% Jianjie Fang$^4$, 
% Chen Gao$^{1}$ \\ 
% Yong Li$^{1}$, 
% Jinqiang Cui$^2$, 
% Xinlei Chen$^{1\ddagger}$, 
% Xiao-Ping Zhang${^1}$
% }

% \affiliation{%
%   \institution{$^1$Tsinghua University,$^2$Pengcheng Laboratory}
%   \institution{$^3$Sun Yat-sen University, $^4$Northeastern University at Qinghuangdao}
% }

\renewcommand{\shortauthors}{Weichen Zhang et al.}

\begin{abstract}
Spatial reasoning is a fundamental capability of multimodal large language models (MLLMs), yet their performance in open aerial environments remains underexplored. In this work, we present Open3D-VQA, a novel benchmark for evaluating MLLMs' ability to reason about complex spatial relationships from an aerial perspective. The benchmark comprises 73k QA pairs spanning 7 general spatial reasoning tasks—multiple-choice, true/false, and short-answer formats—and supports both visual and point cloud modalities. The questions are automatically generated from spatial relations extracted from both real-world and simulated aerial scenes. Evaluation on 13 popular MLLMs reveals that: 1) Models are generally better at answering questions about relative spatial relations than absolute distances, 2) 3D LLMs fail to demonstrate significant advantages over 2D LLMs, and 3) Fine-tuning solely on the simulated dataset can significantly improve the model's spatial reasoning performance in real-world scenarios. We release our benchmark, data generation pipeline, and evaluation toolkit to support further research: \url{https://github.com/EmbodiedCity/Open3D-VQA.code}. 
\end{abstract}

% \begin{teaserfigure}
%   \includegraphics[width=\textwidth]{sampleteaser}
%   \caption{Seattle Mariners at Spring Training, 2010.}
%   \Description{Enjoying the baseball game from the third-base
%   seats. Ichiro Suzuki preparing to bat.}
%   \label{fig:teaser}
% \end{teaserfigure}
\maketitle

\vspace{-0.5cm}
\section{Introduction}

\begin{figure*}[t!]
  \centering
  \includegraphics[width=1.0\linewidth]{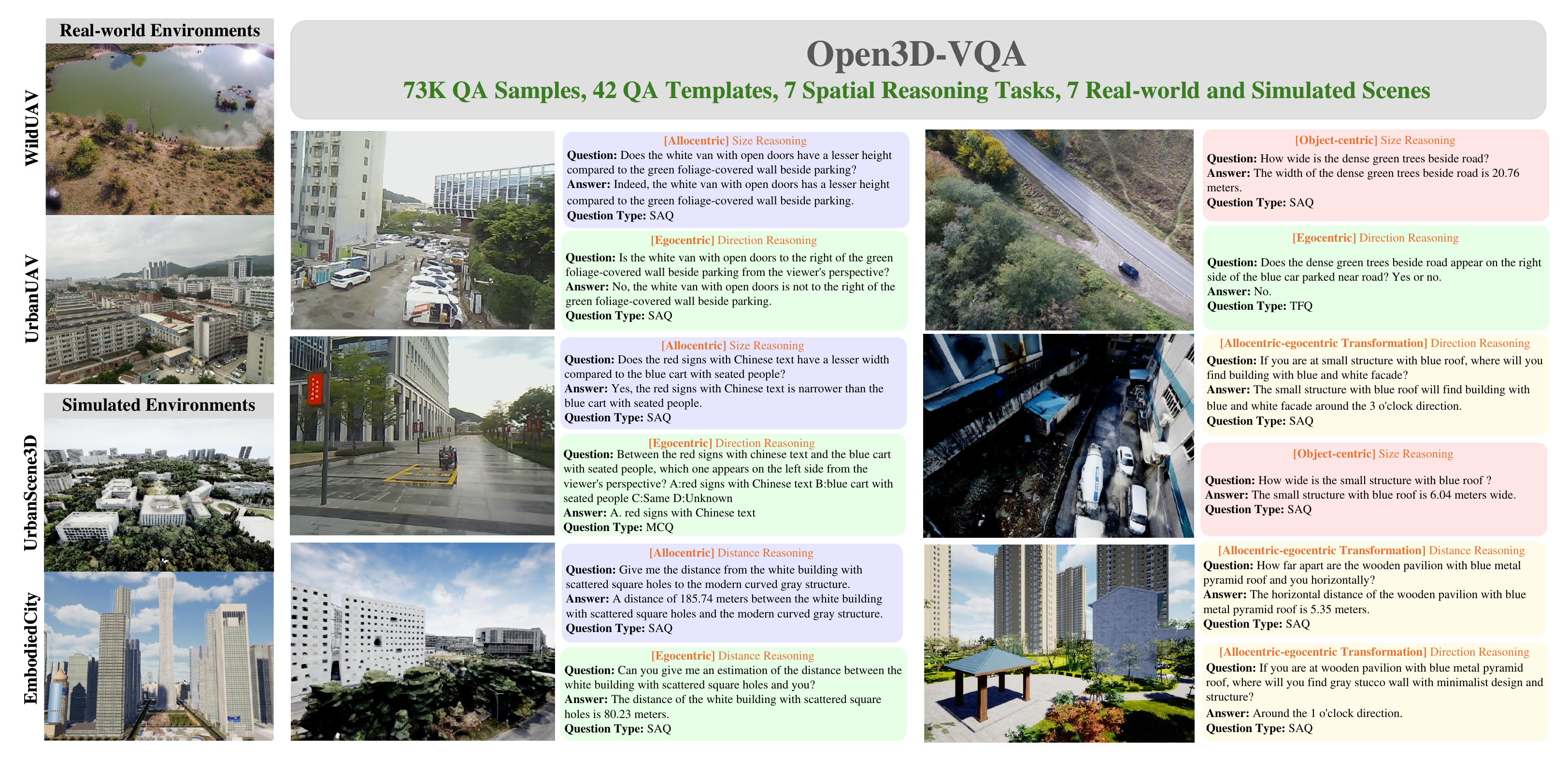}
  \vspace{-0.5cm}
  \caption{\textbf{The overview of Open3D-VQA.} This work includes integration of real-world and simulated data collection platforms, an automatic toolchain for QA generation, and a large-scale aerial spatial reasoning benchmark.}
  \vspace{-0.4cm}
  \label{fig:cover}
\end{figure*}

A fundamental objective within the field of AI research is to equip intelligent agents with the ability to understand spatial information in complex three-dimensional environments, which is essential for various embodied tasks, including vision-and-language navigation \cite{gadre2023cows, majumdar2022zson,liu2023aerialvln}, robotic manipulation \cite{huang2024rekep, driess2022learning}, situation reasoning \cite{linghu2024multi, man2024situational}, and more. 
However, existing question-answering (QA) benchmarks used to evaluate these capabilities are often limited to object-object spatial relationships, lacking the spatial relationship between the object and the agent \cite{liu2023visual, cheng2024spatialrgpt, chen2024spatialvlm}. Situated QA benchmarks \cite{ma2022sqa3d, shiri2024empirical} have considered spatial relationships from the egocentric perspective. However, they focus solely on relative spatial relations and overlook the agent’s ability to perceive precise measurements such as distance. Moreover, these benchmarks are constructed from ground-level perspectives within constrained indoor environments. Thus, the 3D spatial reasoning abilities for urban open-ended spaces have not been well-defined or evaluated. Spatial reasoning in urban spaces possesses the following characteristics:

\begin{itemize}[leftmargin=8pt]
    \item \textbf{Complex Urban Semantics}: Urban scenes encompass complex city layouts, multi-level structures, and open-vocabulary object distributions, posing the challenges for spatial comprehension and reasoning. 
    \item \textbf{Large-scale Spatial Perception}: Unlike indoor environments where agents typically perceive objects within a 10-meter range \cite{yang2024thinking}, urban environments span vast, open areas requiring agents to perceive and reason over much larger distances.
    \item \textbf{Diverse 3D Viewpoints}: Spatial reasoning in urban spaces involves not only ground-level but also aerial perspectives, introducing unique reasoning logic. For example, in an oblique aerial view, buildings situated lower in the field of view may appear closer to the drone.
\end{itemize}

These characteristics introduce new challenges to spatial reasoning in 3D urban environments, and we believe that evaluating this capability offers insights for spatial intelligence~\cite{zhao2025urbanvideo} and urban applications~\cite{chen2024ddl, chen2024soscheduler}.

However, constructing such a benchmark is far from trivial. The difficulties lie in three folds: 1) Designing a comprehensive spatial QA benchmark: The questions must cover a wide spectrum of diverse urban spatial relationships while aligning with natural human language usage. 2) Diverse-perspective aerial data collection: Unlike existing aerial-view datasets such as VisDrone~\cite{cao2021visdrone}, our goal is to capture UAV observations from varying altitudes and camera tilt angles. It requires drones to navigate through dense urban environments, facing risks such as signal loss and potential collisions, making the data collection process risky and costly. 3) Extracting accurate 3D spatial relationships: Generating spatial QA pairs requires a precise understanding of object-level 3D relationships in the scene. Extracting such information requires depth maps, camera intrinsics/extrinsics, and UAV trajectories. However, obtaining these modalities typically demands additional onboard sensors such as depth cameras or RTK, leading to increased costs and extra labor.

In this work, we introduce \textbf{Open3D-VQA}, a novel benchmark for spatial reasoning in 3D urban environments. First, we systematically define three primary types of spatial reasoning tasks and four distinct spatial perspectives. By analyzing the characteristics of each reasoning type under different perspectives, we identify seven distinct spatial reasoning tasks that capture key features of urban spatial understanding. For each task, we design corresponding multiple-choice, true/false, and open-ended question formats, as illustrated in Figure~\ref{fig:cover}.
Second, to collect data from diverse viewpoints, we collaborate with experienced UAV pilots to fly drones across both real-world urban areas and high-fidelity digital twin environments, such as EmbodiedCity~\cite{gao2024embodiedcity} and UrbanScene3D~\cite{lin2022capturing}. We further augment our dataset by incorporating open-source outdoor UAV datasets like WildUAV~\cite{florea2021wilduav}, enhancing both the scene and viewpoint diversity.
Third, we develop a fully automated QA generation pipeline that leverages off-the-shelf models to infer 3D spatial relationships from single RGB images and generates linguistically coherent questions through carefully designed templates. Besides, we introduce a multi-modal correction flow that incorporates ground-truth data from multiple modalities (e.g., depth, camera pose) to reduce the prediction error accumulation through the pipeline.
Finally, we conduct both qualitative and quantitative evaluations of popular MLLMs, including visual and point cloud modalities. We further apply supervised fine-tuning (SFT) on classic models, Qwen~\cite{wang2024qwen2-vl} and LLaVA~\cite{liu2024visual}, to validate the effectiveness and applicability of our proposed Open3D-VQA benchmark.

Our main contributions are:
% \vspace{-0.2cm}
\begin{itemize}[leftmargin=8pt]
    \item We propose \textbf{Open3D-VQA}, a novel question-answering benchmark designed for spatial reasoning in 3D urban environments. The benchmark encompasses four distinct spatial perspectives and seven task types, providing a comprehensive evaluation of an embodied agent’s 3D spatial reasoning capabilities.
    \item We introduce a scalable QA generation pipeline that extracts 3D spatial relationships and generates diverse QA formats from a single RGB image. We design a plug-and-play multi-modal correction flow that leverages available ground-truth information across modalities to reduce error accumulation and ensure high-quality QAs.
    \item We evaluate mainstream MLLMs on Open3D-VQA, revealing their current limitations in spatial reasoning and analyzing their sim-to-real capacities.
\end{itemize}

\section{Related Works}

\subsection{Benchmark for Spatial Reasoning}
Recent advancements in large language models have demonstrated impressive common-sense reasoning abilities across a wide range of tasks, such as task planning \cite{shridhar2020alfred, wang2022self}, navigation \cite{liu2023aerialvln, anderson2018vision, krantz2020beyond}, and manipulation. With the integration of multimodal inputs (e.g., images, point clouds), there has been an increasing focus on evaluating the spatial reasoning capabilities of these models. Prior benchmarks focus on four main reasoning categories: (1) relative spatial reasoning (e.g., CLEVR~\cite{johnson2017clevr}, VSR~\cite{liu2023visual}), (2) absolute spatial reasoning (e.g., SpatialVLM~\cite{chen2024spatialvlm}, SpatialRGPT~\cite{cheng2024spatialrgpt}), (3) situational reasoning involving agent-object relations (e.g., Spatial-MM~\cite{shiri2024empirical}, DriveMLLM~\cite{guo2024drivemllm}), and (4) object-centric reasoning (e.g., GPT4Point~\cite{qi2024gpt4point}, PointLLM~\cite{xu2025pointllm}). However, existing efforts typically cover only a subset of these categories.

To address this, we propose Open3D-VQA, a unified benchmark for 3D spatial reasoning in aerial space that integrates all four VQA types and supports both RGB and point cloud data. This enables a comprehensive evaluation of MLLMs’ spatial reasoning abilities. A comparison with prior benchmarks is provided in Table~\ref{tab::comparison}.

\subsection{Spatial Reasoning via MLLMs}
Predicting the spatial relationships between objects in environments is a fundamental spatial cognition ability of humans. Tons of vision-language models (VLMs)~\cite{lu2024deepseek,lv2023kosmos,team2023gemini,yao2024minicpm,wang2024qwen2-vl,wang2023cogvlm,hurst2024gpt} integrate visual and textual inputs to directly infer spatial relationships. However, due to the absence of spatial measurements, VLMs struggle to predict spatial relations such as distance and length. Other works \cite{hong20233d,zhu20233d,man2024lexicon3d,jia2025sceneverse,xu2025pointllm} have incorporated depth maps or point clouds to provide spatial measurement information, enabling more accurate spatial reasoning. 
% For example, SpatialRGPT designs a plugin module for depth input, significantly improving the spatial reasoning capacity of VLMs by pre-training on a self-generated dataset. Additionally, GPT4Point\cite{qi2024gpt4point} focuses on object point clouds to reason about the spatial attributes of objects.

However, previous works have only covered a subset of spatial VQA tasks in their benchmarks, which has led to an incomplete evaluation of the spatial reasoning capacities of MLLMs.

% \vspace{-1em}  % 缩小表格上方间距
\setlength{\tabcolsep}{4pt}  % 缩小列间距
\renewcommand{\arraystretch}{1.1}  % 控制行高
\begin{table*}[t!]
    \caption{\textbf{Comparisons of our Open3D-VQA with other spatial reasoning benchmarks.} \textit{Qual.}, \textit{Quan.}, \textit{Situ.}, and \textit{Obj.} denote qualitative, quantitative, situational, and object-centric QA, respectively.}
    \vspace{-0.2cm}
    \label{tab::comparison}
    \centering
    \scriptsize
    \resizebox{0.85\linewidth}{!}{
    \begin{tabular}{lcccc|cccc|c}
        \toprule
        ~ & Source & Environment & Modality & Perspective  & Qual. & Quan. & Situ.& Obj. & \# of QA \\
        \midrule
        ScanQA\cite{azuma2022scanqa} & Real. & Indoor & RGB+Point Cloud & Ground  & \ding{52} & \ding{56} & \ding{56} & \ding{52} & 41.3k\\
        SQA3D\cite{ma2022sqa3d} & Real.  & Indoor & RGB+Point Cloud & Ground  & \ding{52} & \ding{56} & \ding{56} & \ding{52} & 33.4k \\
        CLEVR \cite{johnson2017clevr} & Sim. & Indoor & RGB & Ground &  \ding{52} & \ding{56} & \ding{56} & \ding{52} & 720k\\
        VSR \cite{liu2023visual} & Real. & Indoor/Outdoor & RGB & Ground  &  \ding{52} & \ding{56} & \ding{56} & \ding{52} & 10.9k \\ 
        Spatial-MM\cite{shiri2024empirical} & Real. & Indoor/Outdoor & RGB & Ground & \ding{52} &  \ding{56} & \ding{52} & \ding{56} & 2.3k \\
        SpatialVLM \cite{chen2024spatialvlm} & Real. & Indoor/Outdoor & RGBD  & Ground & \ding{56} & \ding{52} &\ding{56} & \ding{56} & N/A  \\
        SpatialRGPT-Bench \cite{cheng2024spatialrgpt} & Real. & Indoor/Outdoor & RGBD  & Ground &  \ding{52} & \ding{52} &  \ding{56} & \ding{56} & 1.5k  \\
         DriveMLLM \cite{guo2024drivemllm} & Real. & Outdoor & RGBD  & Ground  & \ding{52} & \ding{52}  & \ding{52} & \ding{56} & 4.6k \\
         EmboidiedCity \cite{gao2024embodiedcity} & Sim. & Outdoor & RGBD & Aerial & \ding{52} & \ding{52} & \ding{52} & \ding{56} & 50.4k \\
        \midrule
        O3DVQA(Ours)  & Real. \& Sim. & Outdoor & RGBD & Aerial &  \ding{52} & \ding{52} & \ding{52} & \ding{52} & 73.3k  \\
        \bottomrule
    \end{tabular}
    \vspace{-0.1cm}
    }
\end{table*}

\begin{figure*}[t!]
  \centering
  \includegraphics[width=0.9\linewidth]{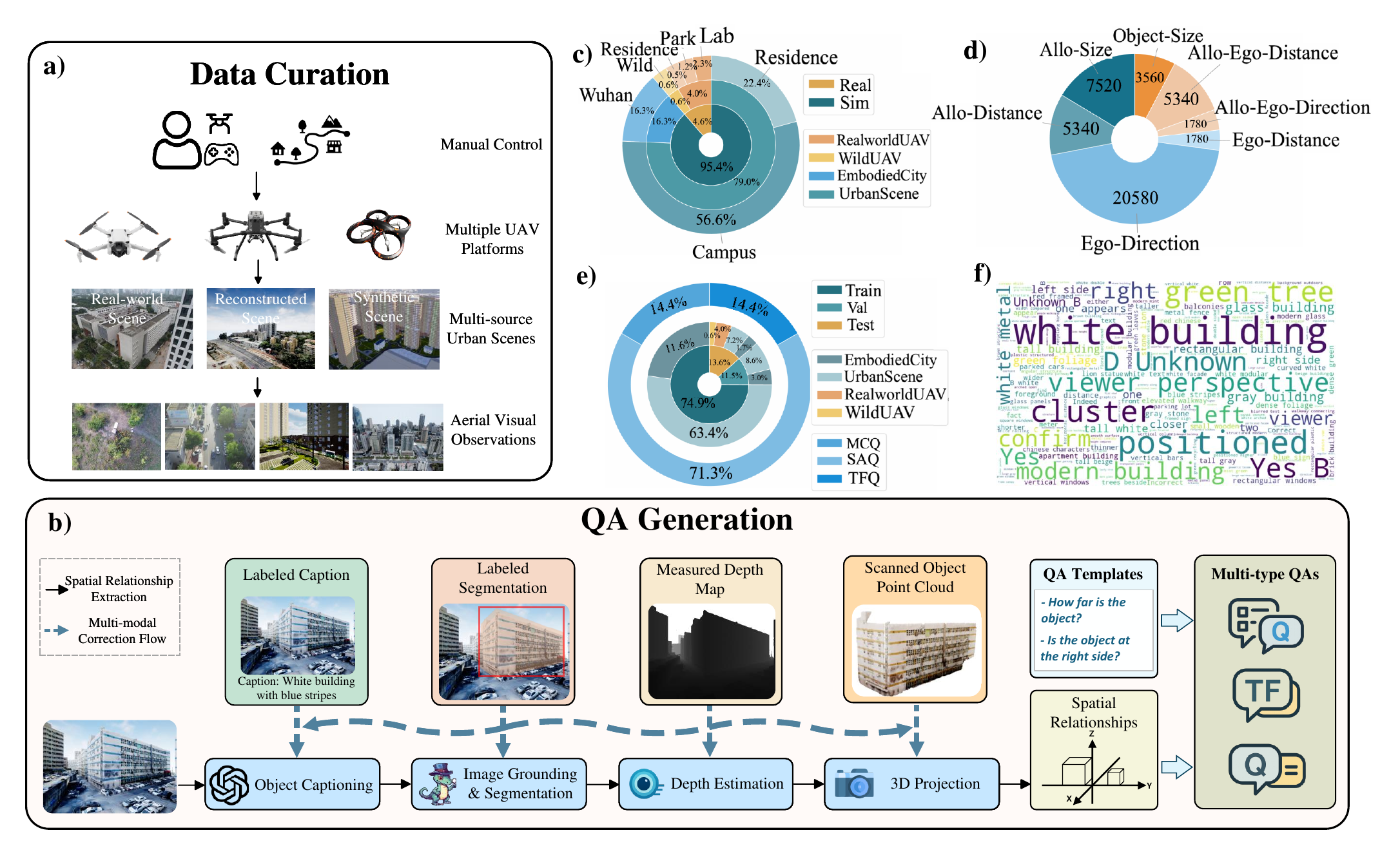}
  \vspace{-0.5cm}
  \caption{\textbf{The data curation pipeline and dataset statistics.}}
  \vspace{-0.4cm}
  \label{fig:data_curation}
\end{figure*}

\section{Benchmark Design and Construction}
% 我们的benchmark定义39种空间QA任务，涵盖了距离、方位、尺寸相关的空间关系，并从egocentric, allocentric, object-centric, allocentric-egocentric transformation 4个方面全面衡量模型的具身空间推理能力。然后，我们介绍了我们的高质量数据采集方法，which is 轻松扩展到其他场景。最后，我们分析了我们数据集的统计属性。

\subsection{Task Set Definition}
To comprehensively evaluate the spatial reasoning capacities of the embodied agent in open space, we categorize spatial reasoning into four distinct types: allocentric spatial reasoning~\cite{shiri2024empirical}, egocentric spatial reasoning~\cite{chen2024spatialvlm}, allocentric-egocentric transformation spatial reasoning~\cite{yang2024thinking}, and object-centric reasoning~\cite {johnson2017clevr}. These categories encompass a diverse range of spatial concepts, including quantitative and qualitative reasoning over distances, orientations, and sizes. We outline the overall task splits in Table~\ref {tab:qa_map}, and detailed tasks are listed in Appendix~\ref{apdx:qa_temp}.

\textbf{Allocentric Spatial Reasoning} evaluates the spatial reasoning capability on object-object relationships from an allocentric perspective, where spatial relationships are independent of the agent's viewpoint or position. Specifically, it includes two types of reasoning tasks: \textit{size reasoning} and \textit{distance reasoning}. The \textit{size reasoning} task requires the agent to infer the relative dimensions, such as comparative length, width, height, and overall size, between pairs of objects. The \textit{distance reasoning} task assesses the agent's ability to reason about the spatial distances between objects, considering the direct distance as well as the horizontal/vertical distance.

\textbf{Egocentric Spatial Reasoning} focuses on spatial relationships between the agent and objects from the agent's perspective, where the relationship depends on the agent's position and orientation. This category includes two specific tasks: \textit{orientation reasoning} and \textit{distance reasoning}. \textit{Orientation reasoning} requires the agent to determine the spatial orientation of an object relative to itself, such as left-right positioning, vertical placement, and angular direction. \textit{Distance reasoning} evaluates the agent's capability to estimate distances between itself and surrounding objects, encompassing direct distance as well as horizontal/vertical distances.

\textbf{Allocentric-egocentric Transformation} evaluates the agent's ability to comprehend spatial relationships across different viewpoints and coordinate systems \cite{yang2024thinking}. Specifically, it assesses the agent's capability to transform spatial information from an allocentric viewpoint into various egocentric perspectives. This transformation involves reasoning about how the orientation relationships between the agent and objects change as the agent moves through the environment. For example, the agent needs to predict how the relative orientation of objects shifts when observing from different viewpoints. Besides, the agent reasons about how the observed distance between objects varies due to the viewpoint change, as distances projected onto different viewpoints can differ significantly.

\textbf{Object-centric Reasoning} focuses on assessing the aerial agent's capability to reason about spatial attributes of urban objects. These attributes, including length, width, height, and overall size, are essential for accurate spatial cognition and effective path planning. Specifically, this reasoning category requires the agent to accurately interpret and quantify these attributes.

\begin{table}[ht]
    \centering
    \small
    \caption{\textbf{Mapping between reasoning capacities and tasks.}}
    \vspace{-0.3cm}
    \resizebox{\linewidth}{!}{
    \begin{tabular}{m{2.5cm}<{\centering}|p{5.5cm}}
    \hline
    \rowcolor{gray!25}
    \textbf{Reasoning Capacity} & \textbf{Reasoning Tasks} \\
    \hline
    
    \rowcolor{blue!10}
    \textbf{Allocentric} & \thead[l]{
    \textbf{Size reasoning}: Infers relative size relationships be-\\tween two objects in space, such as longer/shorter,\\ wider/narrower, taller/shorter, larger/smaller. \\
    \textbf{Distance reasoning}: Infers straight-line, vertical, or \\ horizontal distances between objects. } \\
    \hline
    
    \rowcolor{green!10}
    \textbf{Egocentric} & \thead[l]{
    \textbf{Direction reasoning}: Infers the direction of an object\\ relative to the agent, such as left, right, up, and down. 
    \\
    \textbf{Distance reasoning}: Infers the straight-line distance\\ of an object from the agent.} \\
    
    \hline
    \rowcolor{yellow!10}
    \textbf{Allocentric-egocentric Transformation} & \thead[l]{
    \textbf{Direction reasoning}: The agent infers the direction\\ of objects relative to itself based on its movement. 
    \\
    \textbf{Distance reasoning}: The agent infers object distance\\ in the horizontal or vertical direction relative to itself.} \\
    \hline 
    
    \rowcolor{red!10}
    \textbf{Object-centric} & \thead[l]{
    \textbf{Size reasoning}: Infers the absolute size of a single\\ object, such as its length, width, or height.} \\
    \hline
    \end{tabular}
}
\vspace{-0.4cm}
\label{tab:qa_map}
\end{table}

\subsection{Benchmark Construction Pipeline}
Our dataset construction pipeline is illustrated in Fig~\ref{fig:data_curation}. The dataset is built upon aerial RGB images captured from diverse UAV platforms across both real-world and synthetic urban environments. We introduce a scalable QA generation pipeline that generates multi-type QAs from a single RGB image without any annotation. Given that multi-source data collected from different platforms and scenes inherently includes additional information such as depth maps and camera poses, we further propose a plug-and-play \textbf{multi-modal correction flow}. This method propagates the available multi-modal ground truth in the QA generation pipeline, effectively reducing accumulated errors and enhancing the QA quality.

\subsubsection{Data Curation}
We have three considerations for our data curation phase. 1) \textbf{Scene Diversity}: To prevent the benchmark from being biased toward specific environments, we collect drone flight images and videos from multiple open-source datasets and simulators, including Urbanscene3D, EmbodiedCity, and WildUAV. Urbanscene3D provides reconstructed urban scenes in the UE4 simulator along with real-world aerial images from six distintive areas in Shenzhen. EmbodiedCity offers high-fidelity digital twin cities of Beijing and Wuhan. WildUAV provides real-world overhead imagery from Romania. In addition, we self-collect real-world drone flight videos in Shenzhen. As a result, our dataset covers 4 distinctive real-world areas and 3 synthetic scenes, totaling 4,675 images. 2) \textbf{Hardware Platform Diversity}: The data are gathered using various UAV platforms to avoid bias toward any particular UAV visual sensor or flight system. Specifically, our benchmark includes data collected using four different UAV platforms, including a DJI M300RTK, a DJI Matrice 210, a self-made UAV, and an AirSim simulator-based drone. 3) \textbf{Viewpoint Diversity}: Images in our benchmark are captured from diverse aerial viewpoints, covering a comprehensive range of UAV poses within open 3D space. WildUAV and UrbanScene3D primarily include nadir (top-down) and oblique views from low altitudes. EmbodiedCity provides front-view images in the simulators. To further enrich viewpoint diversity, we manually control UAVs in simulators to traverse various altitudes from low to high and capture multi-view images.

\subsubsection{QA Generation Pipeline}
The key information required for generating QA is the spatial relationships between objects within the scene, which depend on accurate object captions and their corresponding 3D locations. To achieve this, we propose a spatial relationship extraction (SRE) pipeline that explicitly grounds objects within the 3D scene, allowing precise extraction of their spatial relationships. Furthermore, we introduce a multi-modal correction flow (MCF) designed to leverage multi-modal ground-truth data to propagate ground truth information throughout the SRE pipeline, mitigating error accumulation and enhancing the accuracy of the generated spatial relationships.

\definecolor{rank1}{RGB}{255,102,102}
\definecolor{rank2}{RGB}{255,153,153}
\definecolor{rank3}{RGB}{255,204,204}
\definecolor{sectiongray}{RGB}{235, 245, 255}
\definecolor{tie1}{gray}{0.6}
\definecolor{tie2}{gray}{0.8}

\renewcommand{\arraystretch}{1.3}
\setlength{\tabcolsep}{3pt}

\begin{table*}[ht]
\centering
\small
\caption{\textbf{Performance of MLLMs across Spatial Reasoning Tasks.} The gray cell indicates the best performance among all models.}
\resizebox{\linewidth}{!}{
\begin{tabular}{
lcc | ccccccc | ccccccc| ccccccc
}
\toprule
& \multicolumn{1}{l}{}                         
& \multicolumn{1}{l|}{}
& \multicolumn{7}{c|}{\textbf{Total}}
& \multicolumn{7}{c|}{\textbf{Real World}}
& \multicolumn{7}{c}{\textbf{Simulator}}
\\

& \multicolumn{1}{l}{}                         
& \multicolumn{1}{l|}{}                         
& \multicolumn{2}{c}{\cellcolor{blue!10}Allo.} 
& \multicolumn{2}{c}{\cellcolor{green!10}Ego.}        
& \multicolumn{2}{c}{\cellcolor{yellow!10}Trans.}             
& \multicolumn{1}{c}{\cellcolor{red!10}Obj.}                      
& \multicolumn{2}{c}{\cellcolor{blue!10}Allo.} 
& \multicolumn{2}{c}{\cellcolor{green!10}Ego.}        
& \multicolumn{2}{c}{\cellcolor{yellow!10}Trans.}             
& \multicolumn{1}{c}{\cellcolor{red!10}Obj.}                       
& \multicolumn{2}{c}{\cellcolor{blue!10}Allo.} 
& \multicolumn{2}{c}{\cellcolor{green!10}Ego.}        
& \multicolumn{2}{c}{\cellcolor{yellow!10}Trans.}             
& \multicolumn{1}{c}{\cellcolor{red!10}Obj.}

\\
    
\textbf{Method} & \textbf{Rank} & \textbf{Avg.} &
\rotatebox{90}{Size} &
\rotatebox{90}{Distance} &
\rotatebox{90}{Direction} &
\rotatebox{90}{Distance} &
\rotatebox{90}{Direction} &
\rotatebox{90}{Distance} &
\rotatebox{90}{Size} &
\rotatebox{90}{Size} &
\rotatebox{90}{Distance} &
\rotatebox{90}{Direction} &
\rotatebox{90}{Distance} &
\rotatebox{90}{Direction} &
\rotatebox{90}{Distance} &
\rotatebox{90}{Size} &
\rotatebox{90}{Size} &
\rotatebox{90}{Distance} &
\rotatebox{90}{Direction} &
\rotatebox{90}{Distance} &
\rotatebox{90}{Direction} &
\rotatebox{90}{Distance} &
\rotatebox{90}{Size} \\
\midrule

\rowcolor{sectiongray}
\multicolumn{24}{l}{\textbf{Proprietary 2D LLMs}} \\
GPT-4o-mini & 5 & 39.8 & 39.2 & 2.5 & 47.5 & 1.7 & 8.9 & 0.9 & 0.6 & 41.8 & 2.9 & 48.1 & 0.0 & 10.0 & 0.0 & 0.0  & 37.8 & 2.5 & 47.1 & 1.8 & 8.8 & 0.9 & 0.6 \\
GPT-4o & {4} & 47.1 & 62.0 & 4.9 & 51.2 & 2.4 & 5.7 & 1.2 & 2.6 & 68.9 & 5.7 & 52.2 & 0.0 & 0.0 & 0.0 & 0.0 & 58.4 & 4.8 & 50.5 & 2.6 & 6.1 & 1.3 & 2.8 \\
Gemini-2.0-Flash & \cellcolor{rank2}2 & 48.6 & 61.3 & 1.2 & 53.9 & 0.6 & 7.3 & 0.0 & 0.3 & 67.1 & 5.7 & 55.6 & 0.0 & 8.3 & 0.0 & 0.0 & 58.2 & 0.8 & 52.8 & 0.6 & 7.3 & 0.0 & 0.3  \\
Gemini-2.5-Flash & \cellcolor{rank1}1 & 51.6 & 59.5 & 2.1 & 58.7 & 0.6 & \cellcolor{tie1}{32.7} & 1.7 & 0.6 & 65.5 & 3.1 & 59.6 & 0.0 & 25.0 & 8.3 & 0.0 & 56.3 & 2.0 & 58.0 & 0.6 & 33.3 & 1.2 & 0.6 \\
Qwen-VL-Max-latest & \cellcolor{rank3}3 & 47.3 & 56.5 & 1.8 & 53.5 & 0.6 & 9.3 & 0.3 & 1.8 & 61.8 & 0.0 & 53.3 & 0.0 & 8.3 & 0.0 & 0.0 & 53.7 & 1.9 & 53.6 & 0.6 & 9.4 & 0.3 & 1.9 \\

\rowcolor{sectiongray}
\multicolumn{24}{l}{\textbf{Open-source 2D LLMs}} \\
InternVL-4B & 5 & 42.6 & 50.9 & 1.4 & 46.9 & 1.3 & 19.1 & 2.4 & 1.3 & 56.5 & 4.4 & 48.0 & 0.0 & 11.1 & 0.0 & 5.3 & 47.9 & 1.2 & 46.2 & 1.4 & 19.7 & 2.6 & 1.0 \\
InternVL-8B & \cellcolor{rank3}3 & 45.1 & 52.1 & 1.7 & 50.1 & 2.0 & 13.1 & 2.7 & 0.7 & 55.2 & 3.7 & 51.7 & 0.0 & 33.3 & 5.0 & 0.0 & 50.5 & 1.5 & 49.0 & 2.1 & 12.1 & 2.5 & 0.7 \\
LLaVA-1.5-7B & 6 & 37.9 & 36.9 & 0.0 & 45.2 & 0.0 & 1.4 & 0.0 & 0.6 & 37.1 & 0.0 & 45.3 & 0.0 & 12.5 & 0.0 & 0.0 & 36.8 & 0.0 & 45.2 & 0.0 & 0.7 & 0.0 & 0.6 \\
LLaVA-1.5-7B (finetuned) & 4 & 43.0 & 52.3 & 1.3 & 48.3 & 0.0 & 8.1 & 0.3 & 0.0 & 54.3 & 2.9 & 49.1 & 0.0 & 0.0 & 0.0 & 0.0 & 51.2 & 1.2 & 47.9 & 0.0 & 8.6 & 0.3 & 0.0 \\
Qwen2-VL-7B & \cellcolor{rank2}2 & 49.4 & 57.9 & 1.3 & 56.3 & 1.1 & 4.8 & 0.0 & 0.9 & 63.1 & 0.0 & 57.2 & 0.0 & 9.1 & 0.0 & 4.2 & 55.1 & 1.4 & 55.7 & 1.2 & 4.5 & 0.0 & 0.6 \\
Qwen2-VL-7B (finetuned) & \cellcolor{rank1}{1} & 64.0 & \cellcolor{tie1}{70.0} & 0.8 & \cellcolor{tie1}{74.3} & 0.0 & 25.4 & 0.3 & 0.0 & 74.0 & 0.0 & 75.6 & 0.0 & 16.7 & 0.0 & 0.0 & 67.8 & 0.8 & 73.4 & 0.0 & 26.1 & 0.3 & 0.0 \\

\rowcolor{sectiongray}
\multicolumn{24}{l}{\textbf{Open-source 3D LLMs}} \\
3D-LLM & \cellcolor{rank1}{1} & 43.8 & 36.0 & \cellcolor{tie1}{22.4} & 49.3 & \cellcolor{tie1}{42.3} & 22.9 & \cellcolor{tie1}{43.6} & \cellcolor{tie1}{20.5} & 48.5 & 41.7 & 50.2 & 75.0 & 66.7 & 58.3 & 0.0 & 28.8 & 20.9 & 48.6 & 39.9 & 19.6 & 42.5 & 21.8 \\
LEO & \cellcolor{rank2}{2} & 43.4 & 49.2 & 3.4 & 49.3 & 0.0 & 11.2 & 1.2 & 1.2 & 49.1 & 2.9 & 51.5 & 0.0 & 12.5 & 0.0 & 0.0 & 49.3 & 3.4 & 47.9 & 0.0 & 11.1 & 1.3 & 1.2 \\

\bottomrule
\end{tabular}
}
\vspace{-0.3cm}
\label{tab:full_model_performance}
\end{table*}

\textbf{Spatial Relationship Extraction} As illustrated in Figure~\ref{fig:data_curation}, the spatial relationship extraction (SRE) pipeline comprises object captioning, image grounding, depth estimation, and 3D projection modules, following a similar approach to~\cite{VQASynth2023}. In the object captioning module, we prompt GPT-4o to describe distinctive objects within the provided image. We limit the output to at most three objects to exclude ambiguous or trivial objects, such as multiple cars with the same color. In the image grounding module, we utilize SegCLIP~\cite{luo2023segclip} and SAM~\cite{ravi2024sam} to generate bounding boxes and precise masks for the captioned objects. SegCLIP, which is an open-vocabulary segmentation model, produces initial coarse masks by aligning object captions with semantically relevant regions in the image. We subsequently prompt SAM using pixels from these coarse regions to refine these coarse masks into fine-grained counterparts. For the depth estimation module, we employ VGGT~\cite{wang2025vggt}, an outdoor monocular depth estimation method, to generate accurate depth maps along with corresponding camera parameters. Leveraging the depth maps, refined object masks, and camera parameters, we project the objects into 3D space to compute their locations and 3D bounding boxes. Finally, based on these 3D representations, we extract detailed spatial relationships among the objects, such as distances, directions, and other spatial attributes.

\textbf{Multi-modal Correction Flow} Each module in the SRE pipeline inevitably introduces estimation errors that accumulate throughout the entire process, resulting in ambiguous object captions and inaccurate spatial localization. To mitigate this issue, we propose a multi-modal correction flow (MCF) that propagates the multi-modal ground truth in the SRE pipeline. MCF supports various forms of multi-modal ground truth, including object captions, bounding boxes, segmentation masks, depth maps, and accurate 3D scans.

MCF has two flow directions: downstream propagation and upstream propagation. In downstream propagation, module outputs are directly replaced by their corresponding ground-truth data, thereby effectively preventing errors from propagating downstream and impacting the accuracy of the final 3D location predictions. For upstream propagation, MCF aims to enhance the quality and distinctiveness of object captions by utilizing precise bounding box information or accurate 3D scans. Specifically, given a ground-truth object bounding box, MCF crops the corresponding region from the image and feeds it to GPT-4o, generating more precise and descriptive captions. When an accurate 3D scan of an object is available, MCF projects the object's 3D bounding box into the 2D image plane using camera parameters, thereby deriving the object's 2D bounding box.

Within our benchmark, real-world scenes like WildUAV provide depth maps and camera parameters. Simulated scenes like EmbodiedCity provide depth maps, camera parameters, and 3D scans. All these multi-modal ground truths are used in MCF.

\textbf{Multi-type QA generation}
Our benchmark includes multiple-choice questions (MCQ)~\cite{du2024embspatial,zhao2025urbanvideo}, true-or-false question (TFQ)~\cite{wang2019superglue}, and short-answer questions (SAQ)~\cite{chen2024spatialvlm,cheng2024spatialrgpt}, allowing for diverse evaluation formats. The QA pairs are primarily auto-generated based on extracted spatial relationships and well-designed question templates. During generation, a template is randomly selected from the corresponding task-specific pool. More details can be found in Appendix~\ref{apdx:qa_temp}.
To ensure the benchmark's quality, all generated QA pairs are manually refined. Thanks to the MCF and the multi-modal ground-truth metadata, the 3D projection process is noise-free. Therefore, human annotators only need to verify the image grounding results, significantly reducing the manual effort. We discard images containing ambiguous object descriptions or imprecise segmentations.

\subsection{Data Analysis}
\label{sec:data_analysis}
The dataset comprises multiple modalities, poses, and target masks, with image resolutions set to 640x480, commonly used by current UAV platforms. To ensure diversity and realism in the dataset, we collected a total of 4,675 images from four real-world scenes and three virtual scenes. After manual refinement, 1,168 high-quality images were retained.

To guarantee the diversity of QA types, we designed 34, 12, and 12 templates to generate SAQs, MCQs, and TFQs, respectively, resulting in 73,324 QA pairs in total. We adopt 80\% of QAs from simulators for training, 10\% for validation, and the remaining 10\% combined with QAs from the real world for sim-to-real testing. We further depict the ratio of QAs in different scenes, the number of QAs of different reasoning tasks, and the ratio of different QA types in Figure~\ref{fig:data_curation}c-e. Finally, we generate a word cloud shown in Figure~\ref{fig:data_curation}f to illustrate the contextual richness of our benchmark.

\section{Experiments}
For the proposed outdoor spatial reasoning tasks, we evaluated the performance of 13 popular MLLMs, including visual and 3D LLMs. We further fine-tuned two widely used open-source models to validate the effectiveness of our benchmark. Additionally, we compared the performance of various large models across different spatial reasoning tasks and analyze their failure reasons.

\subsection{Experimental Setups}
\subsubsection{Evaluation metrics}
For MCQs and TFQs, we directly calculate the accuracy of each reasoning task. For SAQs, we follow the evaluation strategies from SpatialVLM~\cite{chen2024spatialvlm}. For questions about relative relationships, such as relative size or orientation, we use GPT-4o to assess the consistency between the model's responses and the ground truth on a binary scale (0 or 1). For questions about absolute measurements, such as object distance or size estimations, GPT-4o extracts numerical values from both the model's response and the ground truth. A response is considered correct if the extracted value falls within the range of 
\([0.75, 1.25]\) relative to the ground truth.

\subsubsection{Implementation Details}
For 3D MLLMs inference, scene point clouds are obtained via the pipeline in Figure~\ref{fig:data_curation} and object point clouds are segmented by the 2D object masks. We also align point clouds with their pretrained coordinate systems. 
2D MLLMs are fine-tuned with LoRA~\cite{hu2021lora} using four NVIDIA H100. As described in Section~\ref{sec:data_analysis}, all models are fine-tuned on simulated QA samples and evaluated on both simulated and real-world QA samples.

\begin{figure}[t!]
  \centering
  \includegraphics[width=0.9\linewidth]{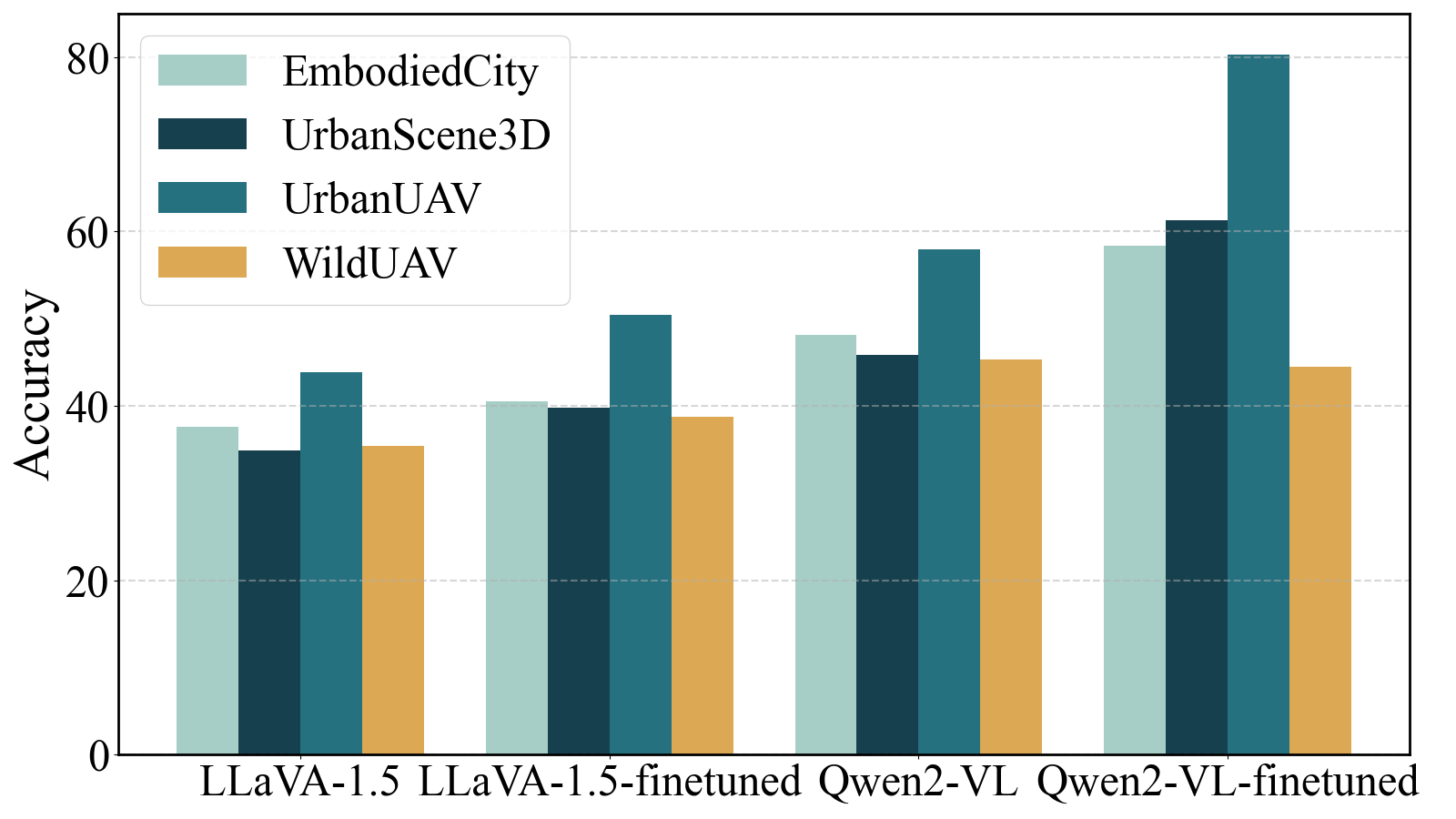}
  \vspace{-0.3cm}
  \caption{\textbf{The average accuracy of LLaVA-1.5 and Qwen2-VL in real-world and simulated scenes.}}
  \vspace{-0.8cm}
  \label{fig:sim2real}
\end{figure}

\subsubsection{Baselines}
We evaluate both 2D and 3D MLLMs. For 2D MLLMs we test both proprietary and open-source models. Proprietary 2D MLLMs include GPT-4o~\cite{achiam2023gpt}, Qwen-VL-Max~\cite{Alibaba2025qwen}, Gemini-2.0 Flash, and Gemini-2.5 Flash~\cite{Google2025gemini}. Open-source 2D MLLMS include InternVL-4B, InterVL-8B~\cite{chen2024internvl}, Qwen2-VL~\cite{wang2024qwen2-vl}, and LLaVA-1.5-7B~\cite{liu2024visual}. For 3D MLLMs, we evaluate 3D-LLM~\cite{hong20233d} and LEO~\cite{huang2023embodied}. 

\subsection{Overall Performance of Baselines}
We present the accuracy of all evaluated models across different spatial reasoning tasks in Table~\ref{tab:full_model_performance}. From these results, we make the following observations and conclusions. 

\textbf{Most models lack allocentric-egocentric transformation reasoning ability.} Compared to other spatial reasoning tasks, all models perform notably worse on reasoning tasks related to allocentric-egocentric transformation. Only Gemini-2.5-Flash, fine-tuned Qwen2-VL, and 3D-LLM achieve accuracy above 20\%, while most models remain below 10\%. The results indicate that current multimodal large language models struggle to shift spatial relationships from an environment-centered view to a self-centered one.

\textbf{Incorporating point cloud information significantly enhances direction and distance reasoning.} All 2D MLLMs perform worse on distance reasoning tasks compared to direction and size reasoning. The average accuracy for distance reasoning across 2D models is only 4.1\%, whereas direction and size reasoning reach 33.2\% and 40.7\%, respectively. The results highlight the challenge current models face in inferring absolute distances. On the other hand, 3D-LLMs achieve comparable performance on distance and direction reasoning, suggesting that point cloud inputs provide models with distance information between objects, thereby strengthening their overall spatial reasoning capability. It is worth noting that although LEO also receives point cloud inputs, its accuracy remains low due to its inability to produce well-formatted responses aligned with the question type, as discussed in Section~\ref{sec:err_analysis}.

% \textbf{Fine-tuning能有效提升模型在方位和尺寸推理任务上的效果}. Both LLaVA-1.5-7B和Qwen2-VL-7B在经过微调之后，在size reasoning和direction reasoning 任务上都有明显的提升。 但是在distance reasoning任务上的行为仍不明显。说明2D MLLM更擅长定性的空间关系推理任务，对于定量的空间关系推理则需要额外的信息，例如深度。
\textbf{Fine-tuning effectively improves model performance on direction and size reasoning tasks.} After fine-tuning, both LLaVA-1.5-7B and Qwen2-VL-7B achieve over 10\% improvements in size reasoning tasks and more than 5\% improvements in direction reasoning tasks, while their performance improvements on distance reasoning remain marginal. This suggests that 2D MLLMs are better suited for qualitative spatial reasoning tasks and require additional spatial information, such as point clouds, for quantitative spatial reasoning.

\begin{figure}[t!]
  \centering
  \includegraphics[width=\linewidth]{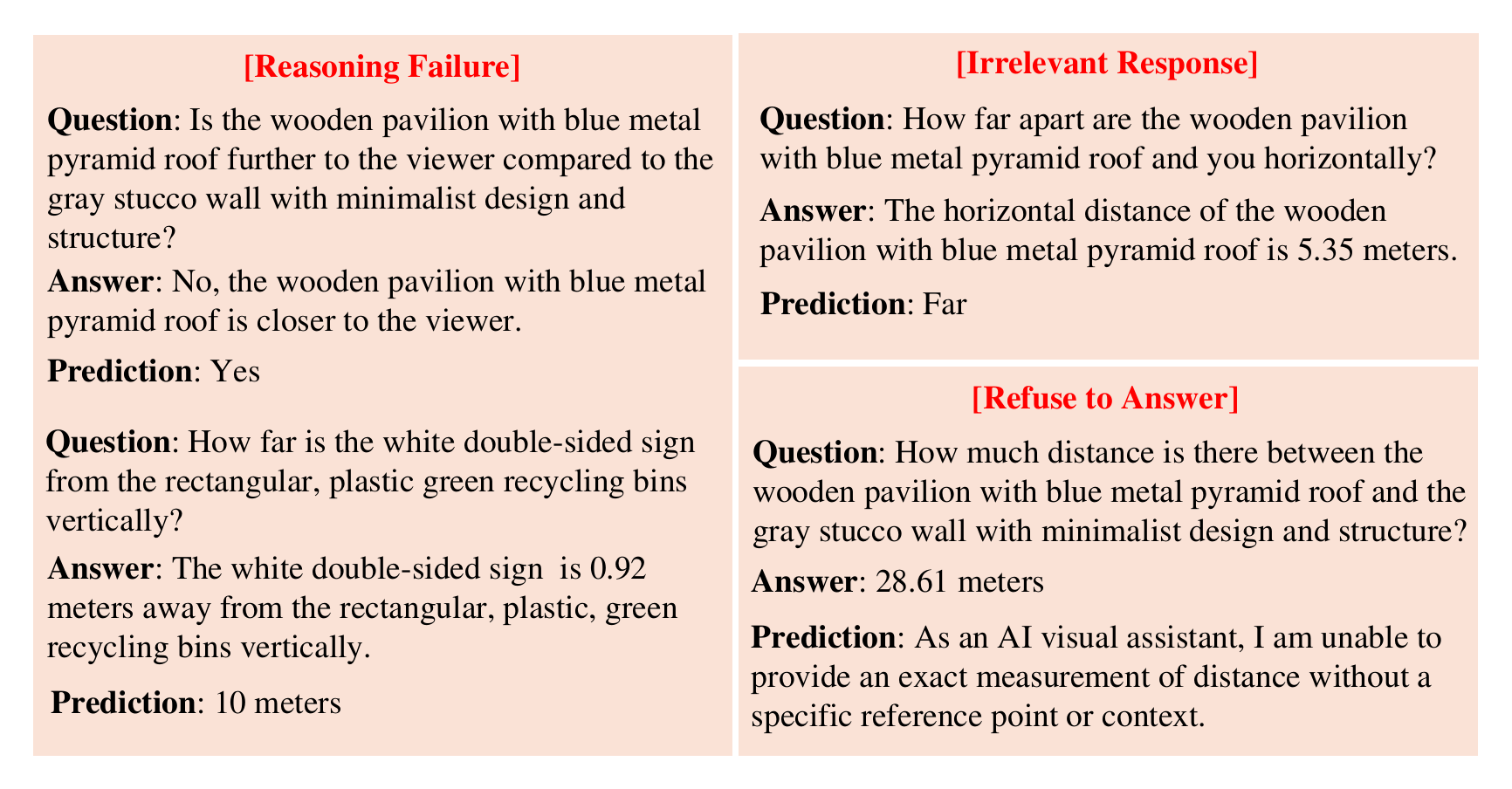}
  \vspace{-0.6cm}
  \caption{\textbf{Three common errors of MLLMs on Open3D-VQA.}}
  \vspace{-0.6cm}
  \label{fig:err_analysis}
\end{figure}

\subsection{Sim-to-real Analysis}

We present the accuracy of the LLaVA-1.5 and Qwen2-VL models under zero-shot and fine-tuning settings across different environments, as shown in Figure~\ref{fig:sim2real}. First, the zero-shot performance on the real-world dataset is comparable to that in simulated environments, indicating that the models possess a certain level of generalization in spatial reasoning. Furthermore, after fine-tuning solely on simulated datasets, the models demonstrate a significant improvement in spatial reasoning performance in real-world scenarios. Specifically, compared to their zero-shot counterparts, LLaVA-1.5 and Qwen2-VL achieve accuracy gains of 6.5\% and 22.3\%, respectively, on the UrbanUAV dataset. This suggests that 2D MLLMs can learn generalizable spatial reasoning capabilities from simulated data and successfully transfer them to real-world environments. These results also validate the effectiveness of our dataset.

\subsection{Failure Analysis}
\label{sec:err_analysis}
% We further analyze the failure cases of MLLMs on spatial reasoning tasks. As depicated in Figure~\ref{fig:err_analysis}, there are three main error sources. The most common one is the reasoning failure that the model failed to give the correct answer of spatial reasoning question. The second failure reason is that the model misunderstands the question and give irrelevant response. 这种现象在LEO模型上尤为严重，也是其虽然能够感知距离信息，但是仍然无法正确回答问题的主要原因。第三种是失败原因是模型拒绝回答问题，如图所示，2D MLLM 如LLaVA在回答距离相关问题时，往往会由于缺少深度信息而选择保守的回答方式。
We further analyze the failure cases of MLLMs on spatial reasoning tasks. As illustrated in Figure~\ref{fig:err_analysis}, there are three primary failure reasons. The most common failure is reasoning errors, where the model is unable to derive the correct answer to a spatial reasoning question despite understanding the input. The second reason for failure is question misinterpretation, where the model fails to comprehend the question and generates an irrelevant response. This issue is especially severe in LEO, which is the major cause of its low accuracy on distance reasoning tasks. The last failure is that models refuse to answer the question. As shown in the figure~\ref{fig:err_analysis}, 2D MLLMs such as LLaVA tend to adopt conservative responses to distance reasoning tasks due to the lack of depth information.

\section{Conclusion}
In this work, we propose Open3D-VQA to comprehensively evaluate the spatial reasoning capacities of both 2D and 3D MLLMs in aerial spaces. We define seven spatial reasoning tasks and design more than 40 QA templates to automatically generate large-scale QAs. 13 popular MLLMs are tested on the benchmark. The results indicate the limited spatial reasoning of MLLMs and the validity of the proposed benchmark. 

\bibliographystyle{ACM-Reference-Format}
\bibliography{sample-base}

\clearpage
\appendix

\section{Appendices}
\subsection{Details of the Data Curation Pipeline}
\subsubsection{Image Caption}
\begin{figure*}[t!]
  \centering
  \includegraphics[width=1.0\linewidth]{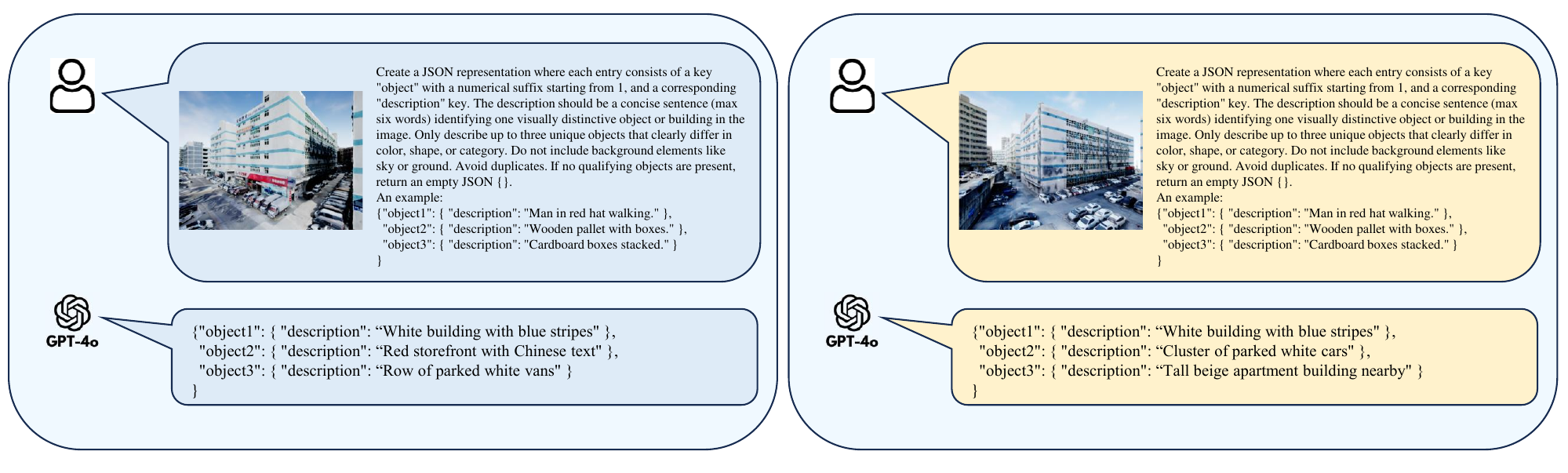}
 %   \vspace{-0.3cm}
  \caption{\textbf{The caption prompt with GPT-4o}}
  % \vspace{-0.4cm}
  \label{fig:caption_prompt}
\end{figure*}

To generate initial captions for the dataset, we utilize GPT-4o by providing it with the RGB image as input. The prompt instructs the model to concisely describe up to three of the most salient objects depicted in the scene. The response is expected in JSON format, where each object is summarized in a short, descriptive phrase. This step serves as the foundation for constructing semantically meaningful scene annotations (see Figure~\ref{fig:caption_prompt} for an example).

\subsubsection{Curation Pipeline}
Given the RGB image and the corresponding object captions produced by GPT-4o, we construct a multi-stage pipeline to generate high-quality visual question-answering (VQA) samples. First, we leverage CLIPSeg to obtain rough semantic segmentations based on the caption keywords, followed by refinement using Segment Anything Model (SAM) to generate precise object masks and bounding boxes. These masks are then projected into 3D space using the aligned depth data, enabling the reconstruction of object-level point clouds for up to three salient objects in the scene.

\begin{figure*}[t!]
  \centering
  \includegraphics[width=1.0\linewidth]{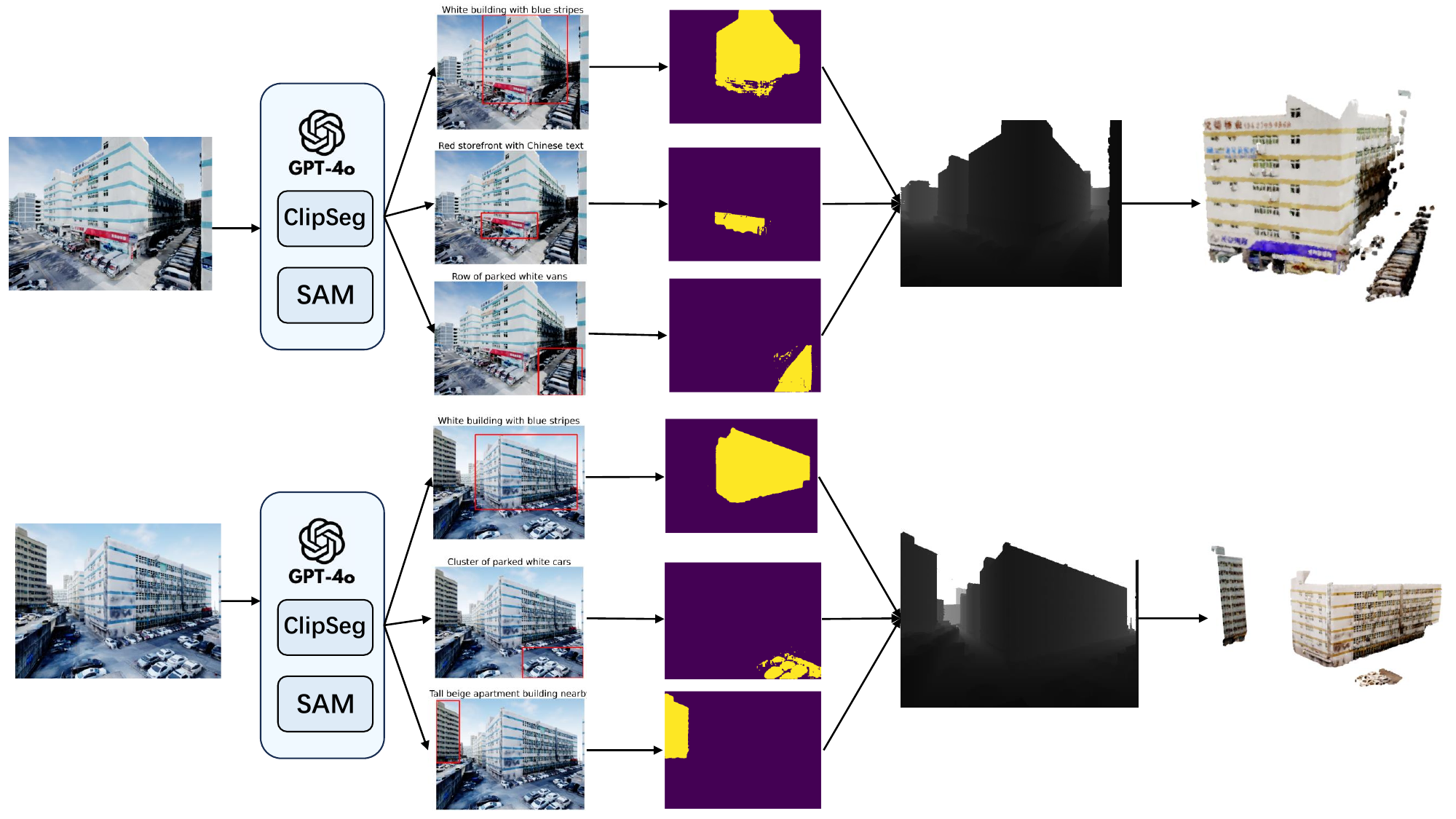}
  \caption{\textbf{Overview of the data curation pipeline.} Starting from RGB and caption inputs, the system performs segmentation, point cloud reconstruction, and QA generation.}
  \label{fig:curation_pipeline}
\end{figure*}

Subsequently, these segmented objects and their spatial relationships provide the basis for generating diverse types of QA pairs. We design a set of templated question generation strategies that cover spatial reasoning, object attributes, and egocentric perspectives. These QA templates are automatically instantiated based on the 3D scene understanding derived from the segmentation and captioning results. See Figure~\ref{fig:curation_pipeline} for an overview of the entire curation pipeline.

\begin{table}[H]
\centering
\begin{tcolorbox}
A chat between a curious human and an artificial intelligence assistant. 

The assistant gives helpful, detailed, and polite answers to the human's questions. 

USER: <image>Is the white carport with pink bolts around to the left of the translucent, modern bus shelter advertisement panel, white frame from the viewer's perspective? 

ASSISTANT: Incorrect, the white carport with pink bolts around is not on the left side of the translucent, modern bus shelter advertisement panel, white frame.
\end{tcolorbox}
\caption{Fine-tuning prompts for LLaVA-1.5}
\label{tab:llava-fine-tune-prompts}
\end{table}

\begin{table}[H]
\centering
\begin{tcolorbox}
SYSTEM: You are a helpful assistant.

USER:

<|image\_pad|><|image\_pad|>...<|image\_pad|> 
Is the white carport with pink bolts around to the left of the translucent, modern bus shelter advertisement panel, white frame from the viewer's perspective?

ASSISTANT:

Incorrect, the white carport with pink bolts around is not on the left side of the translucent, modern bus shelter advertisement panel, white frame.
\end{tcolorbox}
\caption{Fine-tuning prompts for Qwen2-VL}
\label{tab:qwen-fine-tune-prompts}
\end{table}

\begin{table}[H]
\centering
\begin{tcolorbox}
You are a helpful assistant designed to output JSON.
You should help me to evaluate the response given the question and the correct answer.

To mark a response, you should output a single integer between 0 and 1.
1 means that the response perfectly matches the answer.
0 means that the response is completely different from the answer.
\end{tcolorbox}
\caption{GPT-4o prompts for qualitative evaluation.}
\label{tab:qualitative-evaluation-prompts}
\end{table}

\begin{table}[H]
\centering
\begin{tcolorbox}
You are a helpful assistant designed to output JSON.
You should help me to evaluate the response given the question and the correct answer.

You need to convert the distance of the correct answer and response to meters. The conversion factors are as follows: 1 inch = 0.0254 meters. 1 foot = 0.3048 meters. 1 centimeter (cm) = 0.01 meters. 
You should output two floats in meters, one for the answer, and one for the response.
The output should be in JSON format.

\end{tcolorbox}
\caption{GPT-4o prompts for quantitative evaluation.}
\label{tab:quantitative-distance-evaluation-prompts}
\end{table}

\begin{table}[H]
\centering
\begin{tcolorbox}

You are a helpful assistant designed to output JSON.
You should help me to evaluate the response given the question and the correct answer.

You need to extract the direction of the correct answer and response. 
You should output two integers in clock directions, one for the answer, and one for the response.

\end{tcolorbox}
\caption{GPT-4o prompts for direction evaluation.}
\label{tab:quantitative-direction-evaluation-prompts}
\end{table}

\subsection{QA templates of Open3D VQA construction}
\label{apdx:qa_temp}

In this section, we present several representative QA templates from each category used to construct our Open3D VQA dataset. Owing to space constraints, we are unable to include the complete set. Code and dataset will be released to public upon publication. The examples are listed in Table \ref{tab:allo_qa_template}, Table \ref{tab:ego_qa_template}, Table \ref{tab:allo_ego_trans_qa_template} and Table \ref{tab:obj_qa_template}.

\begin{table*}[ht]
    \centering
    \caption{Allocentric QA templates}
    \resizebox{1.0\textwidth}{!}{
    \begin{tabular}{c|c}
    \hline
    \rowcolor{gray!25} \textbf{Templates Name} & \textbf{Example} \\
    \hline
    \rowcolor{blue!10}
    \multicolumn{2}{l}{\textbf{Size reasoning}: Relative size relationships between two objects in space, such as length, width, height, or overall size.} \\
    \hline
    tall\_predicate &  Can you confirm if the [A] is taller than the [B]? \\ \hline
    short\_predicate & Can you confirm if the [A] is shorter than the [B]? \\ \hline
    width\_predicate & Does the [A] have a greater width compared to the [B]?  \\ \hline
    thin\_predicate &  Is the [A] thinner than the [B]? \\ \hline
    tall\_multichoice & Who is taller, the [A] or the [B]? A:[A] B:[B] C:Same D:Unknown  \\   \hline
    short\_multichoice &   Between the [A] and the [B], which one has less height? A:[A] B:[B] C:Same D:Unknown \\ \hline
    wide\_multichoice & Which of these two, the [A] or the [B], appears wider? A:[A] B:[B] C:Same D:Unknown \\ \hline
    thin\_multichoice & Who is thinner, the [A] or the [B]? A:[A] B:[B] C:Same D:Unknown \\ \hline
    big\_tfqa & Does the [A] have a larger size compared to the [B]? A.Yes. B.No. \\ \hline
    small\_tfqa & Does the [A] have a larger size compared to the [B]? A.Yes. B.No. \\ \hline
    wide\_tfqa & Is the [A] wider than the [B]? A.Yes. B.No. \\ \hline
    thin\_tfqa & Can you confirm if the [A] is thinner than the [B]? A.Yes. B.No. \\ \hline
    tall\_tfqa & Is the [A] taller than the [B]? A.Yes. B.No. \\ \hline
    short\_tfqa & Does the [A] have a lesser height compared to the [B]? A.Yes. B.No. \\ \hline
    
    \rowcolor{blue!10}
    \multicolumn{2}{l}{\textbf{Distance reasoning}: The distance between objects along different spatial axes, such as straight-line or axis-aligned distances.} \\
    \hline
    distance\_data & Could you measure the distance between the [A] and the [B]? \\ \hline
    vertical\_distance\_data & What is the vertical distance between the [A] and the [B]? \\ \hline
    horizontal\_distance\_data & Can you give me an estimation of the horizontal distance between the [A] and the [B]? \\ \hline
    \end{tabular}
}
\label{tab:allo_qa_template}
\end{table*}

\begin{table*}[ht]
    \centering
    \caption{Egocentric QA templates}
    \resizebox{1.0\textwidth}{!}{
    \begin{tabular}{c|c}
    \hline
    \rowcolor{gray!25} \textbf{Templates Name} & \textbf{Example} \\
    \hline
    \rowcolor{green!10}
    \multicolumn{2}{l}{\textbf{Direction reasoning}: The object's position relative to the \textbf{agent}, such as left/right, above/below, or angle.} \\
    \hline
    left\_predicate &  Is the [A] to the left of the [B] from the viewer's perspective? \\ \hline
    right\_predicate & Does the [A] appear on the right side of the [B]? \\ \hline
    above\_predicate & Can you confirm if the [A] is positioned above the [B]?  \\ \hline
    below\_predicate &  Can you confirm if the [A] is positioned below the [B]? \\ \hline
    front\_predicate &  Is the [A] in front of the [B]?
    \\ \hline
    behind\_predicate &  Is the [A] positioned behind the [B]? \\ \hline
    
    left\_multichoice & Which is more to the left, the [A] or the [B]? A:[A] B:[B] C:Same D:Unknown \\   \hline
    right\_multichoice &  \makecell{Between the [A] and the [B], which one appears on the right side from the viewer's perspective? \\ A:[A] B:[B] C:Same D:Unknown} \\ \hline
    above\_multichoice & Who is higher up, the [A] or the [B]? A:[A] B:[B] C:Same D:Unknown \\ \hline
    below\_multichoice & Which is below, the [A] or the [B]? A:[A] B:[B] C:Same D:Unknown \\ \hline
    front\_multichoice & \makecell{Between the [A] and the [B], which one appears on closer from the viewer's perspective? \\ A:[A] B:[B] C:Same D:Unknown} \\ \hline
    behind\_multichoice & Who is positioned further to viewer, the [A] or the [B]? A:[A] B:[B] C:Same D:Unknown \\ \hline
    
    left\_tfqa & Is the [A] to the left of the [B] from the viewer's perspective? A.Yes. B.No. \\ \hline
    right\_tfqa & Does the [A] appear on the right side of the [B]? A.Yes. B.No. \\ \hline
    above\_tfqa & Can you confirm if the [A] is positioned above the [B]? A.Yes. B.No. \\ \hline
    below\_tfqa & Is the [A] below the [B]? A.Yes. B.No. \\ \hline
    front\_tfqa & Does the [A] come in front of the [B]? A.Yes. B.No. \\ \hline
    behind\_tfqa & Does the [A] lie behind the [B]? A.Yes. B.No \\ \hline
    
    left\_relation2agent & Is the [A] to the left of you from the viewer's perspective? \\ \hline
    right\_relation2agent & Does the [A] appear on the right side of you? \\ \hline
    above\_relation2agent & Can you confirm if the [A] is positioned above you? \\ \hline
    below\_relation2agent & Does the [A] appear under? \\ \hline
    direction2agent & Estimate the direction of [A]. \\ \hline
    
    \rowcolor{green!10}
    \multicolumn{2}{l}{\textbf{Distance reasoning}: The distance from the object to the \textbf{agent}.} \\
    \hline
    distance2agent & Could you provide the distance between the [A] and you? \\ \hline
    \end{tabular}
}
\label{tab:ego_qa_template}
\end{table*}

\begin{table*}[ht]
    \centering
    \caption{Allocentric-egocentric Transformation QA templates}
    \resizebox{1.0\textwidth}{!}{
    \begin{tabular}{c|c}
    \hline
    \rowcolor{gray!25} \textbf{Templates Name} & \textbf{Example} \\
    \hline
    \rowcolor{yellow!10}
    \multicolumn{2}{l}{\textbf{Direction reasoning}: The angular relation between objects from the agent’s perspective.} \\
    \hline
    direction\_data & If you are at [A], where will you find [B]? \\ \hline
    
    \rowcolor{yellow!10}
    \multicolumn{2}{l}{\textbf{Distance reasoning}: The horizontal or vertical distance between objects from the agent’s perspective along different coordinate axes.} \\
    \hline
    vertical\_distance2agent & How far is the [A] from you vertically? \\ \hline
    horizontal\_distance2agent & Measure the horizontal distance from the [A] to you. \\ \hline
    \end{tabular}
}
\label{tab:allo_ego_trans_qa_template}
\end{table*}

\begin{table*}[ht]
    \centering
    \caption{Objcentric QA templates}
    \resizebox{1.0\textwidth}{!}{
    \begin{tabular}{c|c}
    \hline
    \rowcolor{gray!25} \textbf{Templates Name} & \textbf{Example} \\
    \hline
    
    \rowcolor{red!10}
    \multicolumn{2}{l}{\textbf{Distance reasoning}: The distance between objects along different spatial axes, such as straight-line or axis-aligned distances.} \\
    \hline
    width\_data & What is the width of the [A]? \\ \hline
    height\_data& What is the approximate height of the [A]? \\ \hline
    \end{tabular}
}
\label{tab:obj_qa_template}
\end{table*}

\subsection{Prompt Design for MLLMs}
In this part, we provide a few examples about the prompt we use during our fine-tuning.

\subsubsection{Prompt for Finetuning}
For LLaVA-1.5 and Qwen2-VL, we use the prompt shown in Table \ref{tab:llava-fine-tune-prompts} and \ref{tab:qwen-fine-tune-prompts}, respectively. Note that the <|image\_pad|><|image\_pad|>...<|image\_pad|> part of Qwen2-VL means the 336 image token created by the processor.

\subsubsection{Prompt for Evaluation}
Evaluating our benchmark presents a unique challenge due to the existence of multiple valid answers expressed in varying units. Although human evaluation is capable of addressing such variability, it is often impractical due to its high cost and time requirements. To enable scalable evaluation, we adopt the approach proposed by \cite{cheng2024spatialrgpt}, utilizing GPT-4 to assess the correctness of model outputs.

For qualitative questions, GPT-4 determines whether the model’s response is semantically consistent with the reference answer, assigning a binary score (0 or 1). For quantitative distance questions and direction questions, GPT-4 extracts numerical values from both the ground-truth and the predicted responses. For distance or object attribute questions, GPT-4 is also asked to standardize them to meters. Then we calculate accuracy and error metrics based on the normalized representation. The detailed prompts are listed in Table \ref{tab:qualitative-evaluation-prompts}, \ref{tab:quantitative-distance-evaluation-prompts} and \ref{tab:quantitative-direction-evaluation-prompts}.

\subsection{Qualitative Results of MLLMs}

\begin{figure*}[t!]
  \centering
  \includegraphics[width=0.9\linewidth]{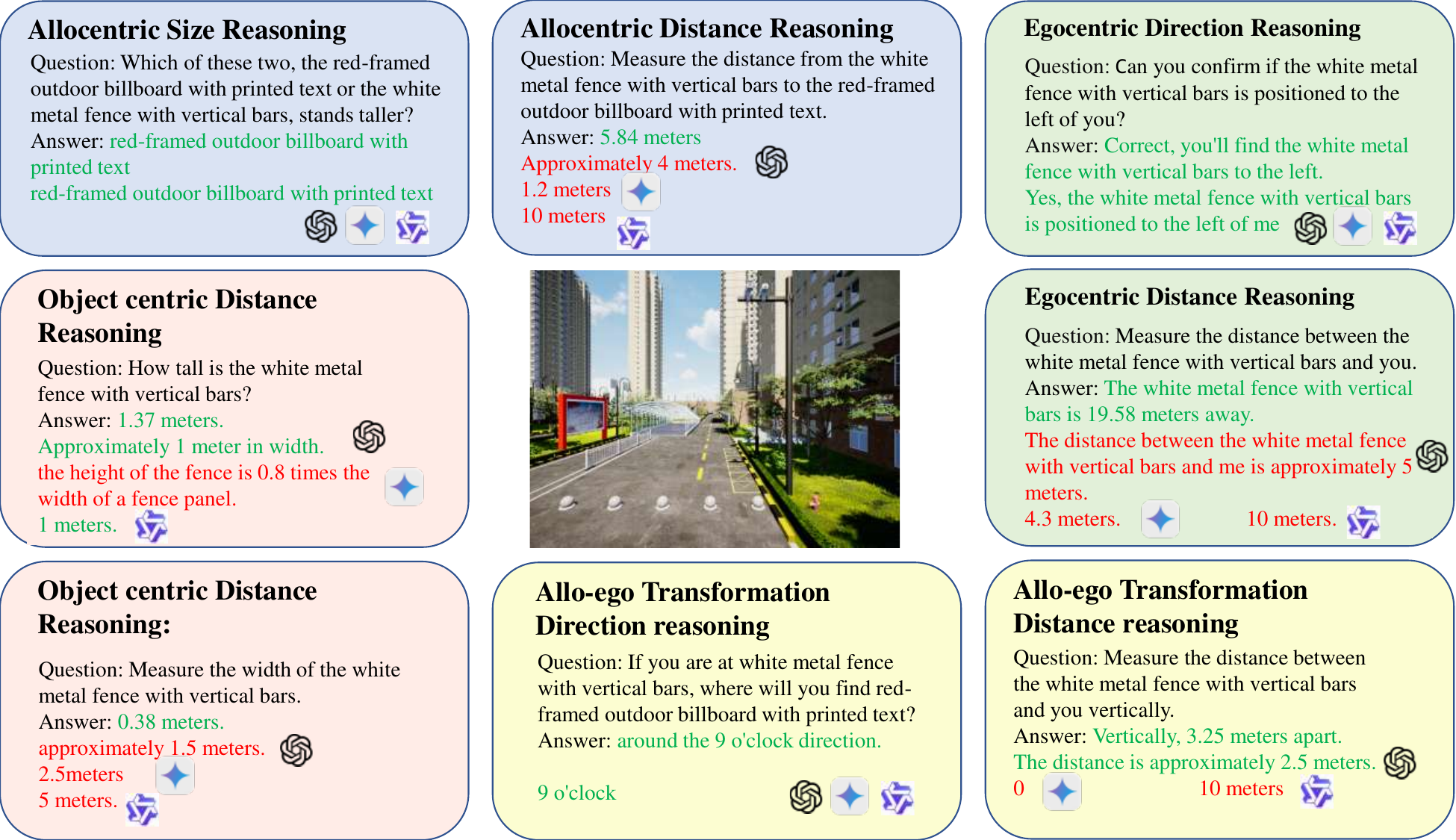}
 %   \vspace{-0.3cm}
  \caption{{Representative Examples of representative model outputs on our benchmark (part I) }}
  % \vspace{-0.4cm}
  \label{fig:model_output_1}
\end{figure*}

\begin{figure*}[t!]
  \centering
  \includegraphics[width=0.9\linewidth]{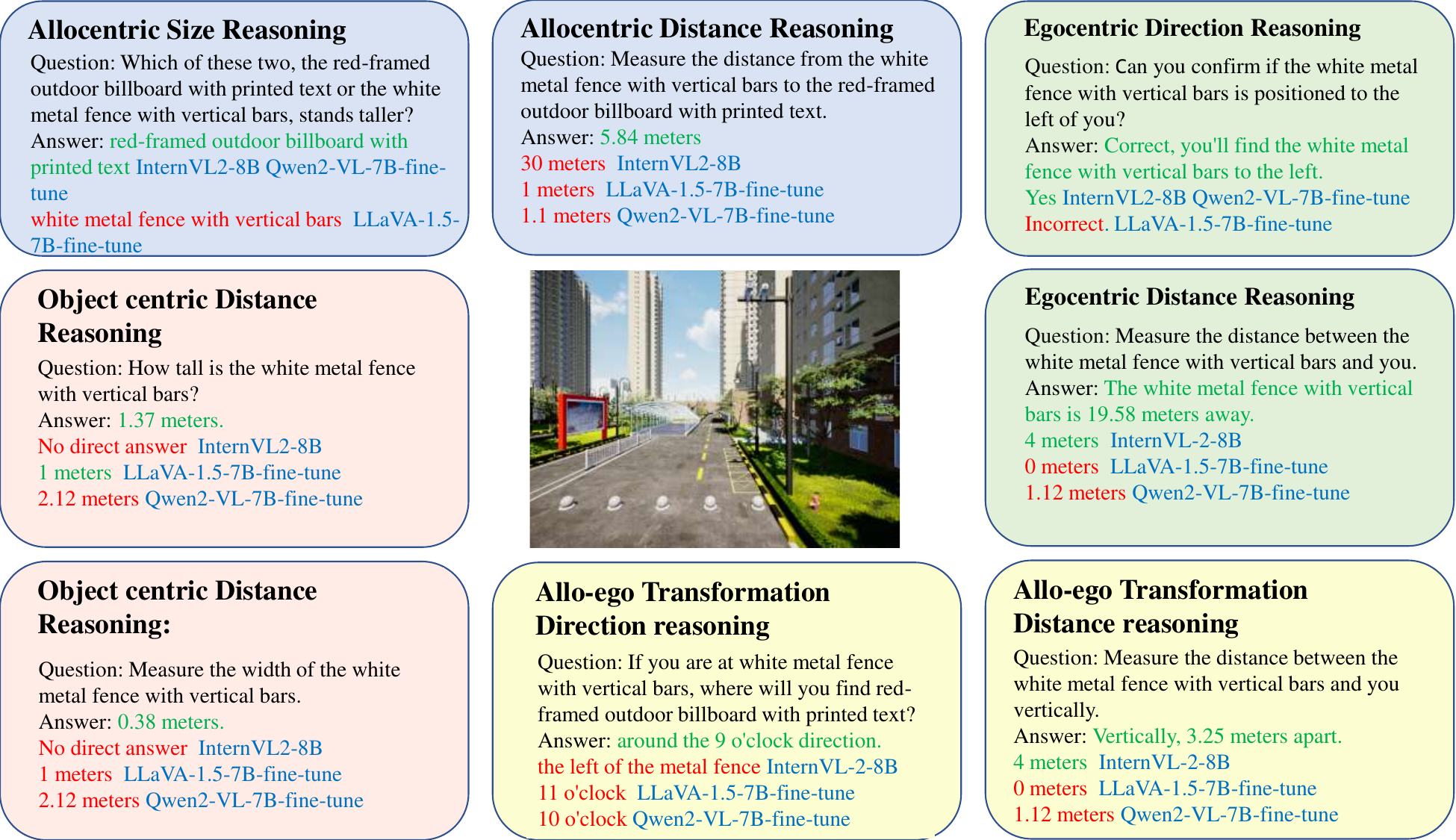}
 %   \vspace{-0.3cm}
  \caption{{Representative Examples of representative model outputs on our benchmark (part II) }}
  % \vspace{-0.4cm}
  \label{fig:model_output_2}
\end{figure*}

\begin{figure*}[t!]
  \centering
  \includegraphics[width=0.9\linewidth]{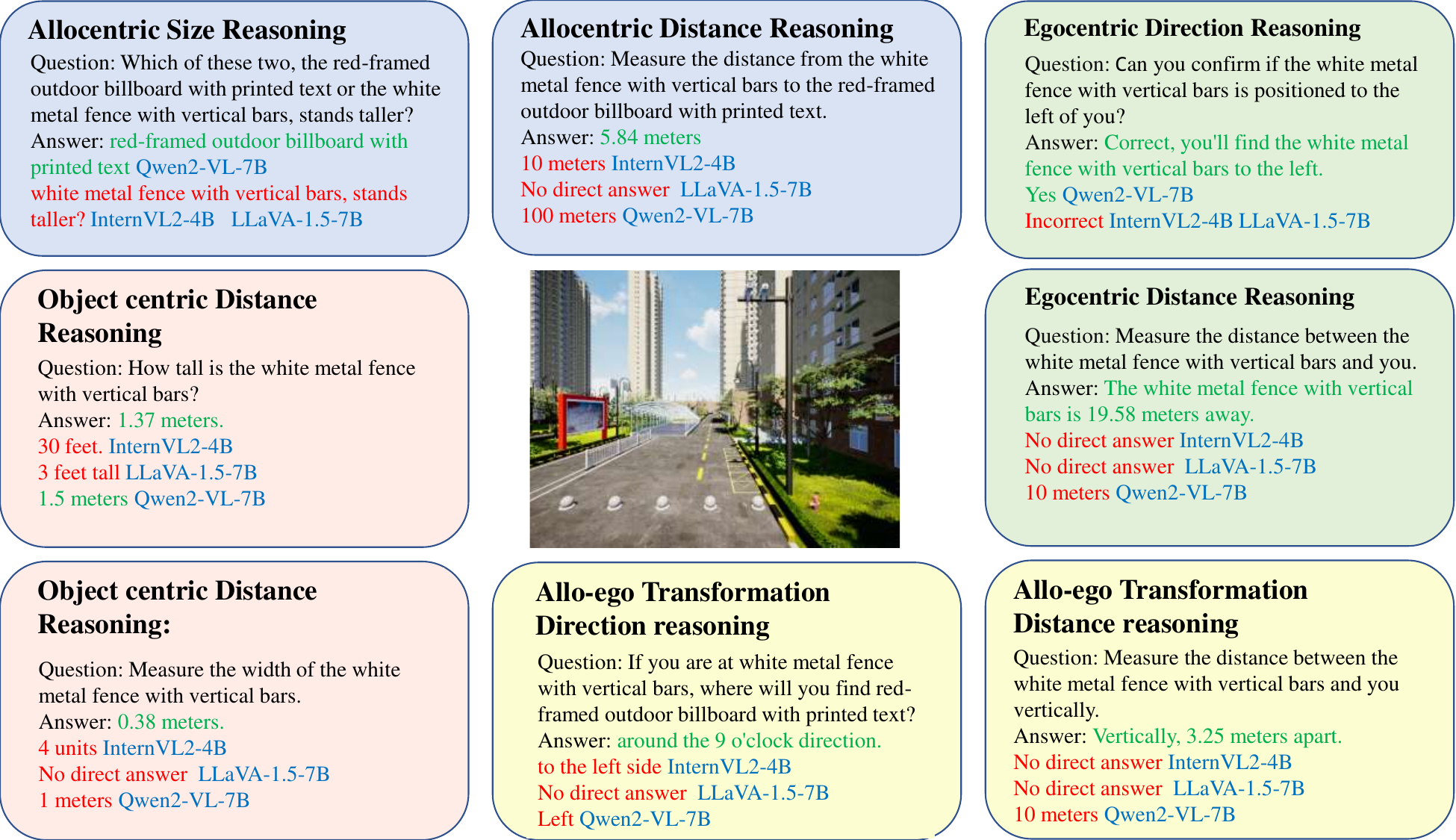}
 %   \vspace{-0.3cm}
  \caption{{Representative Examples of representative model outputs on our benchmark (part III)}}
  \vspace{-0.4cm}
  \label{fig:model_output_3}
\end{figure*}

\begin{figure*}[t!]
  \centering
  \includegraphics[width=0.9\linewidth]{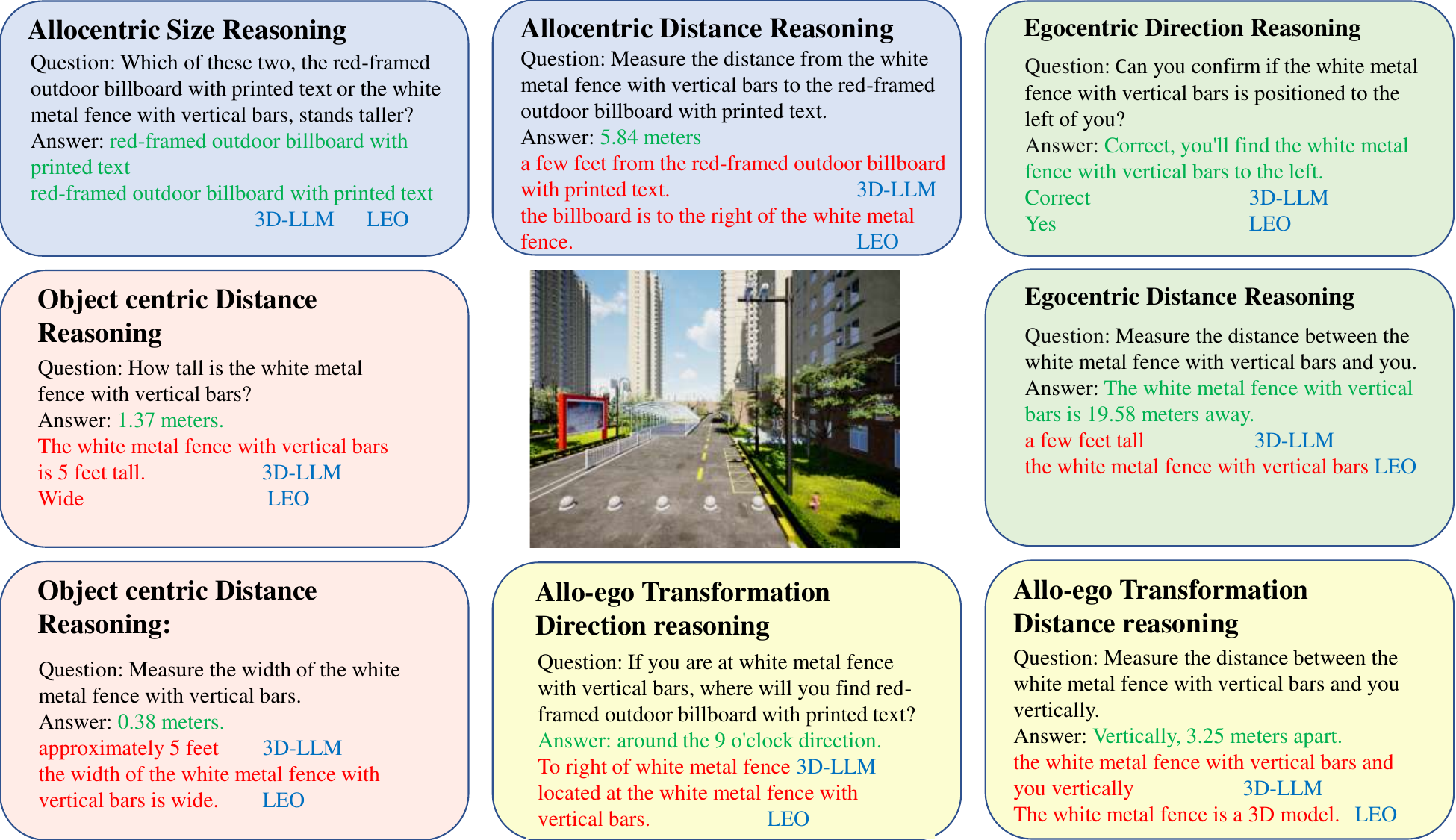}
 %   \vspace{-0.3cm}
  \caption{{Representative Examples of representative model outputs on our benchmark (part IV) }}
  \vspace{-0.4cm}
  \label{fig:model_output_4}
\end{figure*}

In this part, we present qualitative examples of outputs generated by different models during the evaluation phase of our benchmark. Such results serves as a critical complement to former experiment, offering direct evidence of models’ strengths, limitations, and cognitive biases in spatial reasoning tasks. The examples are depicted in Figure \ref{fig:model_output_1}, Figure \ref{fig:model_output_2}, Figure \ref{fig:model_output_3} and Figure \ref{fig:model_output_4}. The names of different models are denoted in blue font. Ground - truth answers and correct responses are presented in green font. Incorrect answers and sections where there is a refusal to answer are indicated in red font.

% In this section, we will present examples of outputs from different models during our evaluation. Examples are shown in Figure \ref{fig:model_output_1}, Figure \ref{fig:model_output_2}, Figure \ref{fig:model_output_3} and Figure \ref{fig:model_output_4}

\vspace{1.0cm}

\end{document}